%% file: main.tex
\documentclass{article}

\PassOptionsToPackage{numbers, compress}{natbib}

\usepackage[preprint]{neurips_2026}

\usepackage[utf8]{inputenc} % allow utf-8 input
\usepackage[T1]{fontenc}    % use 8-bit T1 fonts
\usepackage{hyperref}       % hyperlinks
\usepackage{url}            % simple URL typesetting
\usepackage{booktabs}       % professional-quality tables
\usepackage{amsfonts}       % blackboard math symbols
\usepackage{amssymb}        % \checkmark
\usepackage{nicefrac}       % compact symbols for 1/2, etc.
\usepackage{microtype}      % microtypography
\usepackage{xcolor}         % colors
\usepackage{enumitem}       % custom itemize environment

\usepackage{amsmath}
\usepackage{multirow}
\usepackage{colortbl}
\usepackage{graphicx}

\newcommand{\sref}[1]{\textsuperscript{\ref{#1}}}
\newcommand{\sblk}{\textsuperscript{ }}

\renewcommand{\theenumi}{\arabic{enumi}}
\newcommand{\enumlabel}[1]{\refstepcounter{enumi}\label{#1}\theenumi)}

\newcommand{\ie}{i.e.\ }
\newcommand{\eg}{e.g.\ }

\DeclareMathOperator{\softplus}{softplus}

\DeclareMathOperator{\MLP}{MLP}

\title{Higher-Order Cell Tracking Transformer}

\author{%
  Jord\~ao Bragantini \quad Ilan Theodoro \quad Lo\"ic A.~Royer \\
  Biohub \\
  San Francisco, CA \\
  \texttt{\{jordao.bragantini, ilan.silva, loic.royer\}@biohub.org} \\
}

\begin{document}

\maketitle

\begin{abstract}
    Reconstructing lineages from live-imaging microscopy requires linking cell detections across time, including through cell divisions. A common approach is to construct a candidate graph and associate cell segmentations (nodes) across frames.
    However, these and other existing methods overlook two structural obstacles in candidate tracking graphs: (i)~cell divisions entangle distinct lineage paths in the node embedding space, and (ii)~edges sharing a node have near-random label agreement, so the candidate-graph topology carries no useful information for graph neural networks to aggregate.
    We propose the \textbf{Higher-Order Cell Tracking Transformer} (HOCT), an edge-centric architecture in which candidate cell links attend to one another under a 3D geometric prior, resolving both issues.
    Evaluated on the Cell Tracking Challenge and a bacteria division benchmark, HOCT achieves state-of-the-art results without deep pre-trained image encoders.
    Moreover, the proposed approach is easier to fine-tune, quickly reducing tracking errors by 59\% with 400 annotations in a human-in-the-loop setting, outperforming LoRA fine-tuning of competing transformer baselines (6.75\% improvement).
\end{abstract}

%----------------------------------------------------------------------
\section{Introduction}
\label{sec:introduction}
%----------------------------------------------------------------------

Since the earliest microscopes~\cite{Hooke:1665:Micrographia}, advances in imaging have been inseparable from advances in biology.
Modern live-imaging platforms~\cite{Keller:2008:LightSheet, Chen:2014:LatticeLSM, Bouchard:2015:SCAPE, Yang:2022:Daxi, Shi:2024:SmartLattice} routinely produce terabytes of volumetric time-lapse data, enabling high-content screening~\cite{Feldman:2019:OPS}, the study of embryonic development~\cite{Lange:2024:Zebrahub}, tissue regeneration~\cite{Ccevrim:2025:CrabLeg}, and more.
A fundamental computational challenge in these applications is \emph{cell tracking}: the reconstruction of cell lineages, capturing trajectories and division events over time.

The current dominant paradigm is \emph{tracking by detection}~\cite{Jaqaman:2008:SingleParticalTracking,Mavska:2023:CTC}, in which cells are first segmented in each frame and subsequently linked into lineages.
Unlike multi-object tracking in natural images~\cite{Dendorfer:2020:MOT20}, cells are visually near-identical and lack persistent appearance cues, impairing re-identification methods~\cite{Wojke:2017:DeepSORT, Pang:2021:QuasiDenseSimLearningMOT} that succeed in pedestrian or vehicle tracking~\cite{Bewley:2016:SORT, Bergmann:2019:TrackingWithoutBellsAndWhistles, Meinhardt:2022:TrackFormer}.
Worse, cells can \emph{divide}, changing their morphology and producing a variable number of objects.
Cell tracking has therefore relied on specialized solutions that exploit spatial information, either explicitly via optical-flow~\cite{Hayashida:2020:MPM, Malin:2022:SparseAnnotTracking, Reme:2024:ByoTrack} or implicitly by feeding appearance and positional features to learned models~\cite{Ben:2022:GraphNNCellTracking, Gallusser:2024:Trackastra}.

Recent learning-based approaches employ neural networks on spatiotemporal graphs.
Trackastra~\cite{Gallusser:2024:Trackastra} uses an encoder-decoder Transformer with multi-dimensional Rotary Position Embeddings (RoPE)~\cite{Su:2024:RoPE, Heo:2024:RoPEVision} to embed each cell detection (node) into a latent space and compute association probabilities from cosine similarity between all nodes in adjacent frames.
\citet{Ben:2022:GraphNNCellTracking} and \citet{Braso:2020:LearningNeuralSolverMOT} formulate tracking as edge classification on spatiotemporal graphs, for cell and pedestrian tracking respectively, \cite{Ben:2022:GraphNNCellTracking} using GNN message passing on the candidate graph~\cite{Rozemberczki:2021:PathfinderNetworksGNN}.              

\begin{figure}[ht]
    \centering
    \includegraphics[width=\textwidth]{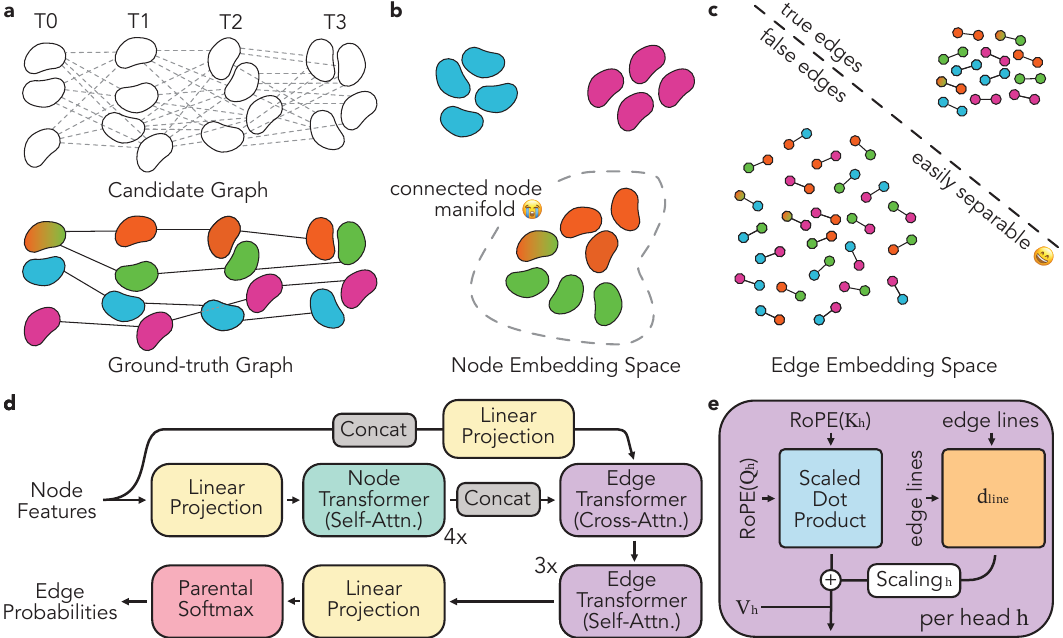}
    \caption{\textbf{Motivation and architecture.}
    (a)~A candidate tracking graph and its ground-truth solution, where each color denotes a distinct cell trajectory (simple path in the lineage tree).
    The graph is non-homophilic: edges sharing a node have near-random label agreement, so graph topology carries no useful label information.
    (b)~In node embedding approaches, divisions create connected manifolds that merge distinct lineage paths, confounding similarity-based association because distinct paths are connected in the embedding space by the division junction.
    (c)~In our edge-centric approach, each candidate link is an independent token; attention is biased by inter-link geometry rather than the candidate-graph topology, removing both the connected-manifold and the non-homophilic topology issues.
    (d)~HOCT architecture: node features are projected and refined through $L_n{=}4$ self-attention layers with 3D RoPE; edge tokens are constructed from node pairs (Eq.~\ref{eq:edge_gather}), then refined through $L_e{=}4$ attention layers with RoPE and line-to-line distance biases; a classification head produces per-edge logits, normalized via the parental softmax.
    (e)~Detail of the edge attention layer with geometric distance bias from edge line segments (Eq.~\ref{eq:attn_dist_bias}).}
    \label{fig:motivation}
    \label{fig:architecture}
\end{figure}

Node-embedding approaches face two structural problems on this graph (Fig.~\ref{fig:motivation}a--b). First, divisions create a \emph{connected-manifold} effect: when a cell $p$ divides into daughters $d_1, d_2$, the embedding of $p$ must be close to both, merging lineage paths that should form separate clusters and leaving them prone to identity swapping. Second, the candidate graph is non-homophilic~\cite{Zheng:2022:GraphHeterophilySurvey, Platonov:2023:CriticalLookHeterophily}: only one (or two, at divisions) of the many edges incident to a target node is correct, so edges sharing a node have near-random label agreement (Sec.~\ref{sec:results}, $\mathcal{H}_{\mathrm{adj}}\approx 0$). Message passing therefore aggregates noise, which explains why edge-classifying GNNs~\cite{Ben:2022:GraphNNCellTracking} have not consistently surpassed baselines.

Both issues are resolved by working in an \emph{edge} embedding space, with attention biased by inter-edge geometry rather than candidate-graph adjacency (Fig.~\ref{fig:motivation}c). We dubbed this approach \textbf{Higher-Order Cell Tracking Transformer} (HOCT), a two-stage architecture where: (1)~node self-attention contextualizes cell detections, and (2)~edge self-attention refines candidate-link representations through edge-to-edge comparisons with geometric distance biases. The term \emph{higher-order} reflects that the model compares edges (second-order structures) to each other. Our main contributions are:
\begin{itemize}[noitemsep, topsep=0pt, parsep=0pt, partopsep=0pt]
    \item An \textbf{edge-centric Transformer architecture} for cell tracking that compares candidate links through geometry-driven self-attention, bypassing the uninformative graph topology and resolving the connected-manifold problem at divisions (Sec.~\ref{sec:architecture}).
    \item A \textbf{geometric distance encoding} based on the line-to-line distance between edge pairs, injected as a learnable, per-head attention bias (Sec.~\ref{sec:distance_encoding}).
    \item A \textbf{multi-frame parental softmax} that extends the single-frame formulation of~\cite{Gallusser:2024:Trackastra} to multiple temporal gaps with an implicit no-parent option, variable appearance costs, and a two-pass tracklet ILP solver (Sec.~\ref{sec:parental_softmax}).
    \item An \textbf{incremental correction} experiment showing that a simple logistic regression head on frozen HOCT features outperforms LoRA fine-tuning of a node transformer~\cite{Lalit:2025:UnsupInteractiveCellTracking}, demonstrating that edge-level representation quality dominates more complex approaches (Sec.~\ref{sec:correction}).
\end{itemize}

%----------------------------------------------------------------------
\section{Related work}
\label{sec:related}
%----------------------------------------------------------------------

\textbf{Cell tracking.}
Classical pipelines formulate tracking as an optimal matching problem on a candidate graph through: linear assignment~\cite{Jaqaman:2008:SingleParticalTracking, Kuhn:1955:HungarianMatching}, network flow~\cite{Zhang:2008:MultiObjTrackingNetworkFlow}, or general integer linear programming (ILP)~\cite{Schiegg:2013:ConservationTracking,Turetken:2016:NetworkFlow, Bragantini:2024:UltrackECCV}, optimizing association costs globally under biological constraints on division, appearance, and disappearance.
Learned matching costs are increasingly popular: optical-flow advection of detections across frames~\cite{Hayashida:2020:MPM, Sugawara:2022:ELEPHANT}, sparse-annotation association costs with ILP-based lineage reconstruction~\cite{Malin:2022:SparseAnnotTracking}, and contrastive cell representations for direct association on centroids~\cite{Zhou:2025:CELLECT} or cell patches~\cite{Han:2025:ASCENT}.

\textbf{Transformers for tracking.}
The Transformer architecture~\cite{Vaswani:2017:Transformer} has been applied to MOT~\cite{Meinhardt:2022:TrackFormer, Zhang:2022:ByteTrack} and object detection~\cite{Carion:2020:DETR}.
Trackastra~\cite{Gallusser:2024:Trackastra} adapts the Transformer for cell tracking with RoPE-modulated attention over node tokens, achieving strong results on the Cell Tracking Challenge~\cite{Mavska:2023:CTC}.

\textbf{Positional encoding.}
Rotary Position Embeddings (RoPE)~\cite{Su:2024:RoPE} encode positions by rotating query and key vectors to modulate the attention between spatially distant tokens. \citet{Heo:2024:RoPEVision} extended RoPE to higher dimensions for vision, and \citet{Liu:2025:RethinkingRoPE} refined it with learnable reflections.
We apply 3D RoPE with learnable per-head frequencies, giving each head an independent spatial encoding.

\textbf{Non-homophilic graphs.}
Most graph neural networks assume \emph{homophily}, the property that connected elements share similar labels, so neighborhood aggregation reinforces correct predictions~\cite{Zheng:2022:GraphHeterophilySurvey}.
When this assumption fails, aggregation mixes uninformative or conflicting signals and degrades representations~\cite{Platonov:2023:CriticalLookHeterophily}; proposed remedies remain topology-driven~\cite{Zheng:2022:GraphHeterophilySurvey}.
Although Transformers can in principle attend to all tokens, spatial positional encodings, such as RoPE, reintroduce a topology-like bias. As in tracking, candidate graphs are constructed by connecting detections within a distance threshold~$\tau$; RoPE decays attention with spatial distance, so the highest-weighted tokens coincide with the graph neighbors.
When attention is further restricted by a spatial mask (as in both Trackastra~\cite{Gallusser:2024:Trackastra} and our node stage), the effective receptive field collapses to the graph neighborhood.
Since candidate tracking graphs are non-homophilic (Sec.~\ref{sec:results}), this amounts to aggregating topologically adjacent but label-uninformative neighbors.
HOCT's edge attention sidesteps this: attention is biased by the line-to-line distance between candidate links, a geometric relationship that captures spatial compatibility beyond the original graph adjacency.

%----------------------------------------------------------------------
\section{Higher-order cell tracking}
\label{sec:method}
%----------------------------------------------------------------------

Let $G_{C} = (V_{C}, E_{C})$ be a candidate graph where $V_{C}$ is the set of cell detections across all frames and $E_{C}$ contains candidate links between detections in different frames within a spatial distance threshold~$\tau$.
Each node $i \in V_C$ carries features $\mathbf{x}_i \in \mathbb{R}^d$, a time dimension $t_i$, and a 3D position $\mathbf{p}_i \in \mathbb{R}^3$.
Each edge $e = (i, j) \in E_C$ connects a source node $i$ at time $t_i$ to a target node $j$ at time $t_j > t_i$, with temporal gap $\Delta t_e = t_j - t_i$.

The goal is to find a solution subgraph $G = (V, E) \subseteq G_C$ that forms a valid cell lineage, \ie a binary forest such that each node has at most one incoming edge (parent) and at most two outgoing edges (allowing for division).
HOCT learns to predict a probability $p_e \in [0,1]$ for each candidate edge $e \in E_C$, and an ILP solver (Sec.~\ref{sec:ilp}) recovers the optimal biologically valid solution.

\subsection{Architecture overview}
\label{sec:architecture}

HOCT is a two-stage Transformer (Fig.~\ref{fig:architecture}d--e) whose attention layers encode spatial coordinates into queries and keys via 3D RoPE~\cite{Su:2024:RoPE} with learnable per-head frequencies and a learnable reflection~\cite{Liu:2025:RethinkingRoPE}.

\textbf{Node self-attention.}
Input features are projected to hidden dimension $C$ via $\mathbf{z}_i^{(0)} = \mathbf{W}_{\mathrm{in}} \mathbf{x}_i$, then refined through $L_n$ self-attention layers.
Each layer applies RoPE to queries and keys using the node position $\mathbf{p}_i$, and masks out node pairs farther than a distance threshold~$\tau$ to preserve sparsity.

\textbf{Edge attention.}
For each candidate edge $e = (i, j)$, we construct an edge token by concatenating the contextualized node features and projecting through an MLP:
\begin{align}
    \mathbf{h}_e^{(0)} = \MLP_{\mathrm{gather}}\bigl([\mathbf{z}_i^{(L_n)} \| \mathbf{z}_j^{(L_n)}]\bigr) \label{eq:edge_gather}
\end{align}
Where $ [ . \| . ] $ is the feature concatenation operator.
The edge position for RoPE is the midpoint $\mathbf{p}_e = (\mathbf{p}_i + \mathbf{p}_j)/2$.
Edge tokens, $\mathbf{h}_e^{(l)}$, are then refined through $L_e$ attention layers, with the first layer performing cross-attention with $\mathbf{f}_e$ as queries and $\mathbf{h}_e^{(0)}$ as keys/values, grounding the edge representation in the raw feature relationships.
Where $\mathbf{f}_e = \mathbf{W}_{\mathrm{edge}}([\mathbf{x}_i \| \mathbf{x}_j])$ are the edge-features tokens computed from the raw input features.
Subsequent layers perform self-attention among edge tokens.
Every edge attention layer incorporates a geometric distance bias (Sec.~\ref{sec:distance_encoding}) based on the line-to-line distance between edge pairs in addition to RoPE.

\textbf{Classification head.}
The final edge embeddings are passed through LayerNorm and a linear projection to produce scalar logits $\ell_e \in \mathbb{R}$, which are normalized via the parental softmax (Sec.~\ref{sec:parental_softmax}).

\subsection{Line-to-line distance encoding}
\label{sec:distance_encoding}

RoPE encodes the position of each token (node or edge midpoint) but ignores the \emph{geometric relationship} between edge pairs.
For instance, two edges can have the same midpoint distance while passing close to each other or running parallel at a distance, encoding distinct motion patterns such as coherent displacement, collision, or position swapping.\footnote{In 3D the line segments rarely strictly intersect; the line-to-line distance still smoothly captures their proximity.}
To provide an explicit geometric signal, we introduce a distance bias based on the 3D line-to-line distance.

Each edge $e = (i, j)$ defines a directed line segment from $\mathbf{p}_i$ to $\mathbf{p}_j$.
For any pair of edges $e_m$ and $e_n$, we compute the minimum Euclidean distance between their line segments:
\begin{align}
    d_{\mathrm{line}}(e_m, e_n) = \min_{s, u \in [0,1]} \left\| \bigl(\mathbf{p}_{i_m} + s \cdot \mathbf{d}_m\bigr) - \bigl(\mathbf{p}_{i_n} + u \cdot \mathbf{d}_n\bigr) \right\| \label{eq:line_dist}
\end{align}
where $\mathbf{d}_m = \mathbf{p}_{j_m} - \mathbf{p}_{i_m}$ is the direction vector of edge $e_m = (i_m, j_m)$.
This distance is computed analytically via the parametric closest-point formula for 3D line segments, with clamping for boundary cases (Appendix~\ref{app:implementation}) and injected into the edge attention logits as a \emph{per-head learnable bias}:
\begin{align}
    A_{h}(m, n) = \frac{\hat{\mathbf{q}}_{h,m}^\top \hat{\mathbf{k}}_{h,n}}{\sqrt{D}} + \sigma_h \cdot \softplus(\alpha_h) \cdot d_{\mathrm{line}}(e_m, e_n) \label{eq:attn_dist_bias}
\end{align}
where $\hat{\mathbf{q}}, \hat{\mathbf{k}}$ are the RoPE-encoded queries and keys, $d_{\mathrm{line}}$ is in dataset spatial units (positions standardized per dataset), $\alpha_h$ is a learnable scalar per head (initialized near zero), and $\sigma_h \in \{+1, -1\}$ is a fixed sign alternating across heads. The \texttt{softplus} prevents the sign from flipping, which we observed to be unstable even with transformations that vary slowly near zero.
With $\sigma_h = -1$, nearby edges receive higher attention (\emph{attractive} heads), encouraging comparison of near-intersecting edges; with $\sigma_h = +1$, distant edges receive higher attention (\emph{repulsive} heads), providing global context about competing trajectories.
Combining both signs yields the best results.

\subsection{Parental softmax and ILP inference}
\label{sec:parental_softmax}
\label{sec:ilp}

A biological constraint in cell tracking is that each cell has at most one parent.
Trackastra~\cite{Gallusser:2024:Trackastra} introduced the \emph{parental softmax} to normalize association scores among candidate parents of the same target node, but applied it only to consecutive frames ($\Delta t = 1$).
We extend it to enable long-range connections, normalizing per target node \emph{and} temporal gap:
\begin{align}
    p_e = \frac{\exp(\ell_e)}{1 + \sum_{e' \in \mathcal{C}(j_e, \Delta t_e)} \exp(\ell_{e'})} \label{eq:parental_softmax}
\end{align}
where $\mathcal{C}(j, \Delta t) = \{e' \in E_C : j_{e'} = j,\; \Delta t_{e'} = \Delta t\}$ is the set of candidate edges targeting node $j$ at temporal gap $\Delta t$.
Normalizing within each $\Delta t$ group ensures that short- and long-range candidates compete only with edges of the same temporal gap, preventing competition between multiple probabilities along the same path.
The constant 1 in the denominator acts as an implicit ``no-parent'' option: when no good match exists, all edge probabilities for that node are driven toward zero.
We train with focal loss~\cite{Lin:2017:FocalLoss} applied to the parental-softmax probabilities, with increased weight on division edges because they are rare and challenging (Appendix~\ref{app:implementation}).

During inference, an ILP solver converts predicted edge probabilities into globally consistent tracks that respect the biological constraints on cell division, appearance, and disappearance, see Appendix~\ref{app:ilp} for full ILP formulation.
The parental softmax also enables an unexplored improvement over standard ILP formulations~\cite{Zhang:2008:MultiObjTrackingNetworkFlow, Turetken:2016:NetworkFlow}, which we describe below.

\textbf{Variable appearance probability.}
Most ILP formulations use a fixed appearance cost~\cite{Gallusser:2024:Trackastra, Bragantini:2025:Ultrack, Bragantini:2024:UltrackECCV, Malin:2022:SparseAnnotTracking}, treating all cells as equally likely to start a new track. This is an unrealistic simplification, for example, new tracks are more likely to appear near the field-of-view boundary. Therefore, we propose using the predicted edge logits to estimate the appearance probabilities.

The constant $1$ in the denominator of Eq.~\eqref{eq:parental_softmax} yields a per-$\Delta t$ ``no-match'' probability $q_{j,\Delta t}$, which we aggregate with a $\lambda$-decayed weighted average over temporal gaps to obtain a per-node appearance cost:
\begin{align}
    q_{j,\Delta t} &= \frac{1}{1 + \sum_{e' \in \mathcal{C}(j, \Delta t)} \exp(\ell_{e'})} \label{eq:q_delta}
\end{align}
Note that when $\Delta t = 1$, $q_{j,1}$ is the probability that node $j$ has no parent in the immediately preceding frame (orphan probability).
The per-temporal gap probabilities $q_{j,\Delta t}$ are then aggregated with a weighted average:
\begin{align}
    p_{\alpha}(j) = \sum_{\Delta t} w_{\Delta t} \cdot q_{j,\Delta t}, \quad
    w_{\Delta t} = \frac{\exp\bigl(-(\Delta t - 1)\lambda\bigr)}{\sum_{\Delta t'} \exp\bigl(-(\Delta t' - 1)\lambda\bigr)}
    \label{eq:appearance}
\end{align}
where $w_{\Delta t}$ gives more weight to shorter temporal gaps.
This reduces the penalty for starting new tracks based on the network's predictions, without any additional loss term.

\textbf{Two-pass tracklet solver.}
Long-range edges ($\Delta t > 1$) bridge missing detections but compete with short-range alternatives along the same path. We resolve this in two passes: (1)~solve the ILP using only $\Delta t = 1$ edges to produce high-confidence tracklets; (2)~aggregate each tracklet into a single meta-node and re-solve over the tracklet graph using all edges. This prevents long-range edges from overriding confident short-range associations.

%----------------------------------------------------------------------
\section{Experiments}
\label{sec:results}
%----------------------------------------------------------------------
\subsection{CTC benchmark}
\label{sec:ctc}

\textbf{Architecture and training.}
HOCT uses $C{=}288$, $H{=}4$ heads, $L_n{=}L_e{=}4$ layers, and distance threshold $\tau{=}300$ pixels.
The model is trained on 16 datasets from the Cell Tracking Challenge (CTC)~\cite{Mavska:2023:CTC}, spanning brightfield, fluorescence, and phase-contrast microscopy in 2D and 3D.
Each dataset contains two annotated time-lapse sequences (32 sequences total); we form two cross-validation splits by assigning one sequence per dataset to each split, training on one and evaluating on the other.
Data augmentation includes spatial crops, affine transformations, and feature dropout that randomly zeroes entire feature groups to encourage robustness to missing features.
Full training details (optimizer, schedule, augmentation, and ILP configuration) are in Appendix~\ref{app:implementation}.

\textbf{Metrics.}
We report standard CTC metrics~\cite{Matula:2015:AOGM, Ulman:2017:ObjectiveCompCTC}:\textbf{LNK} (Linking accuracy); \textbf{BIO} (Biological metric); and \textbf{CLB} (Cell Linking Benchmark), the main ranking metric for the challenge and the average between LNK and BIO;  all in $[0,1]$ with higher being better.
For the bacteria benchmark, we extract scores from~\cite{Achard:2024:TrackastraImgFeatures} and report: \textbf{AOGM} (Acyclic Oriented Graph Metric, lower is better) with its decomposition into addition (AOGM-A) and deletion (AOGM-D) errors, and \textbf{Division~F1}.

\textbf{Topological label information.}
To quantify the non-homophilic claim from Sec.~\ref{sec:introduction}, we build the line graph $L(G_C)$ of each candidate tracking graph (where original edges become nodes that are connected when they share an endpoint) and measure label agreement on it; per-sequence statistics are in Appendix~\ref{app:heterophily}.
Across the CTC sequences, the adjusted homophily~\cite{Platonov:2023:CriticalLookHeterophily} on $L(G_C)$ is $\mathcal{H}_{\mathrm{adj}} = 0.01 \pm 0.04$, indistinguishable from random; for each true (GT) edge, only $29\%$ of its co-incident edges are also true, so $71\%$ of a correct link's line-graph neighbors carry a conflicting label.
For reference, standard homophilic benchmarks have $\mathcal{H}_{\mathrm{adj}} > 0.5$~\cite{Platonov:2023:CriticalLookHeterophily}, where graph neural networks succeed because topology carries the labels; candidate tracking graphs sit at the random baseline.
This explains why GNN-based edge classification~\cite{Ben:2022:GraphNNCellTracking} has not surpassed simpler baselines: message passing aggregates over a topology that is uninformative by construction.

\subsection{Ablations}
\label{sec:ablations}

\textbf{ILP solver components.}
\label{sec:ablation_ilp}
Table~\ref{tab:ablation} ablates the two ILP contributions on the CTC benchmark across two cross-validation splits, both with and without long-range candidate edges ($\Delta t > 1$).
The variable appearance probability is consistently beneficial, improving CLB in every paired comparison (rows 1$\to$2, 3$\to$4, 5$\to$6).
Long links alone degrade performance ($-3.2\%$ CLB, $-5.5\%$ BIO at row 1$\to$3) due to competing edges along the same path but become beneficial when paired with the two-pass tracklet solver, with the full configuration reaching the best CLB ($0.920$) and BIO ($0.858$).

\begin{table}[ht]
  \caption{
    Ablation of ILP solver components on the CTC benchmark.
    Long links require the two-pass tracklet solver: without it, candidate edges of different lengths along the same path compete in the ILP.
    Best results in bold, near-best underlined.
  }
  \label{tab:ablation}
  \centering
  \setlength{\tabcolsep}{3pt}
  \begin{tabular}{@{}ccc ccc ccc ccc@{}}
    \toprule
    \multicolumn{3}{c}{Setup} & \multicolumn{3}{c}{Average} & \multicolumn{3}{c}{Split 1} & \multicolumn{3}{c}{Split 2} \\
    \cmidrule(lr){1-3} \cmidrule(lr){4-6} \cmidrule(lr){7-9} \cmidrule(lr){10-12}
    Long links & 2-pass       & Appear.\ Prob. & CLB & LNK & BIO & CLB & LNK & BIO & CLB & LNK & BIO \\
    \midrule
                 &              &              & 0.909 & \textbf{0.982} & 0.835 & 0.930 & \textbf{0.988} & 0.871 & 0.888 & \underline{0.976} & 0.799 \\
                 &              & $\checkmark$ & 0.914 & \textbf{0.982} & 0.845 & 0.932 & \textbf{0.988} & 0.876 & 0.896 & \textbf{0.977} & 0.815 \\
    $\checkmark$ &              &              & 0.877 & 0.974 & 0.780 & 0.904 & 0.982 & 0.827 & 0.849 & 0.965 & 0.733 \\
    $\checkmark$ &              & $\checkmark$ & 0.897 & 0.977 & 0.816 & 0.920 & 0.984 & 0.857 & 0.873 & 0.971 & 0.775 \\
    $\checkmark$ & $\checkmark$ &              & 0.916 & \underline{0.981} & 0.851 & \underline{0.938} & \underline{0.987} & 0.888 & 0.894 & 0.975 & 0.813 \\
    $\checkmark$ & $\checkmark$ & $\checkmark$ & \textbf{0.920} & \underline{0.981} & \textbf{0.858} & \textbf{0.939} & \underline{0.987} & \textbf{0.890} & \textbf{0.900} & \underline{0.976} & \textbf{0.825} \\
    \bottomrule
  \end{tabular}
\end{table}

\textbf{Edge-stage aggregation mechanism.}
\label{sec:gnn_ablation}
Given the line-graph heterophily measured above, any aggregator restricted to candidate-graph topology (GNN message passing or topology-restricted attention) should combine each edge with label-uncorrelated neighbours.
We test this by replacing HOCT's edge-attention stage with three alternatives, holding everything else fixed (input features, Stage~1, edge gatherer, candidate graph, focal loss, training schedule, and ILP post-processing of Table~\ref{tab:ablation}, last row); GNN variants use eight edge-stage layers to match the transformer in parameter count.

\begin{table}[ht]
  \caption{
  Edge-stage aggregation mechanism ablation: HOCT's edge-attention stage is replaced with three alternatives (Edge-Transformer: global self-attention without RoPE or distance bias; GAT~\cite{Velickovic:2018:GAT}: graph attention on $L(G_C)$; FAGCN~\cite{Bo:2021:FAGCN}: heterophily-aware GNN), holding Stage~1, edge gatherer, candidate graph, loss, training schedule, and ILP post-processing fixed.
  Means and standard deviations across two cross-validation splits.}
  \label{tab:gnn_ablation}
  \centering
  \begin{tabular}{lccc}
    \toprule
    Method & CLB & LNK & BIO \\
    \midrule
    FAGCN & $0.909 \pm 0.003$ & $0.981 \pm 0.001$ & $0.837 \pm 0.004$ \\
    GAT   & $0.916 \pm 0.001$ & $0.982 \pm 0.001$ & $0.851 \pm 0.001$ \\
    Edge-Transformer & $0.922 \pm 0.002$ & $0.982 \pm 0.001$ & $0.862 \pm 0.003$ \\
    HOCT  & $\mathbf{0.926 \pm 0.007}$ & $\mathbf{0.983 \pm 0.001}$ & $\mathbf{0.867 \pm 0.013}$ \\
    \bottomrule
  \end{tabular}
\end{table}

Table~\ref{tab:gnn_ablation} compares~\footnote{\cite{Ben:2022:GraphNNCellTracking} implementation did not allow training on all CTC datasets and therefore was not included}:
(i) Edge-Transformer, a geometry-free analogue of HOCT: global self-attention on edge tokens without the line-to-line distance bias;
(ii) GAT~\cite{Velickovic:2018:GAT}, vanilla graph attention on the line graph $L(G_C)$, the closest analogue to topology-driven GNN tracking~\cite{Ben:2022:GraphNNCellTracking};
(iii) FAGCN~\cite{Bo:2021:FAGCN}, frequency-adaptive sign-aware aggregation as a heterophily-aware GNN baseline.

All three alternatives underperform HOCT on every metric, confirming the heterophily prediction.
The two GNN variants are bounded above by their input topology: aggregating over $L(G_C)$ pulls each true edge's representation toward predominantly incorrect labels even when the layer is sign-aware (FAGCN), consistent with the prediction that the underlying topology carries no label signal for a topology-driven aggregator, attractive or repulsive, to exploit.
The Edge-Transformer baseline, which removes only the geometric biases of Sec.~\ref{sec:distance_encoding} while keeping global attention, also underperforms, isolating the contribution of line-to-line distance encoding from that of unrestricted attention.

\subsection{Official CTC submission}
\label{sec:ctc_submission}

We submitted HOCT to the Cell Tracking Challenge~\cite{Mavska:2023:CTC}; Table~\ref{tab:ctc} reports the official scores computed by the organizers on the \emph{hidden} test sequences,\footnote{The CTC organizers have not yet refreshed the public leaderboard as of 2026-05-01; the scores reported here were provided to us directly.} providing an unbiased comparison against all currently submitted methods (full per-dataset scores and ranks, including the datasets where HOCT falls outside the top three, are listed in Appendix~\ref{app:ctc_full}).
All entries come from a \emph{single} HOCT model trained jointly on all 2D and 3D CTC datasets, testing generalization across diverse cell types and imaging modalities.
This single model ranks first in CLB, LNK, and BIO by overall generalizability and reaches the top three on $14$ of $16$ datasets in at least one metric. Competing methods referenced in the table are
\enumlabel{mt:epfl}~Trackastra~\cite{Gallusser:2024:Trackastra},
\enumlabel{mt:kth}~Baxter Algorithm~\cite{Magnusson:2016:SegmentationNTracking},
\enumlabel{mt:luh}~MAMHT~\cite{Kaiser:2025:MitosisAwareMultiHypothesisTracker},
\enumlabel{mt:mon}~TrackTour~\cite{tracktour},
\enumlabel{mt:past}~ByoTrack~\cite{Reme:2024:ByoTrack},
\enumlabel{mt:rwth}~Medical SAM2 and SAM 3D~\cite{Zhu:2024:MedicalSAM2, Wang:2025:SAMMed3D},
\enumlabel{mt:siat}~zTrack~\cite{zTrack4CTC}, and
\enumlabel{mt:uvic}~MAGIK~\cite{Pineda:2023:GeometricDLMicroMotion}.

% helpers
\begin{table}[t]
\caption{\textbf{Cell Tracking Challenge leaderboard (hidden test set).}
Official scores from the CTC organizers on the held-out test sequences for the three primary metrics (CLB, LNK, BIO; higher is better).
We list the top three submissions per dataset and metric; \textbf{HOCT (ours)} is bolded when in the top three, see text for the full list of competing methods.
The \emph{Average} row reports the official CTC overall-performance (Generalizability) score, the headline cross-dataset measure.}
\label{tab:ctc}
\centering
\small
\begin{tabular}{@{}lccc@{\hspace{8pt}}ccc@{\hspace{8pt}}ccc@{}}
\toprule
& \multicolumn{3}{c}{CLB} & \multicolumn{3}{c}{LNK} & \multicolumn{3}{c}{BIO} \\
\cmidrule(lr){2-4} \cmidrule(lr){5-7} \cmidrule(lr){8-10}
Dataset                    & 1st & 2nd & 3rd & 1st & 2nd & 3rd & 1st & 2nd & 3rd \\
\midrule
Average                    & \textbf{0.930}~\sblk & 0.923~\sref{mt:past} & 0.886~\sref{mt:epfl}
                           & \textbf{0.985}~\sblk & 0.984~\sref{mt:past} & 0.977~\sref{mt:epfl}
                           & \textbf{0.875}~\sblk & 0.862~\sref{mt:past} & 0.794~\sref{mt:epfl} \\
\midrule
HSC                        & 0.873~\sref{mt:past} & \textbf{0.845}~\sblk & 0.807~\sref{mt:luh}
                           & \textbf{0.992}~\sblk & 0.991~\sref{mt:past} & 0.986~\sref{mt:mon}
                           & 0.756~\sref{mt:past} & \textbf{0.699}~\sblk & 0.635~\sref{mt:luh} \\
MuSC                       & \textbf{0.902}~\sblk & 0.894~\sref{mt:luh}  & 0.889~\sref{mt:past}
                           & \textbf{0.986}~\sblk & 0.985~\sref{mt:mon}  & 0.985~\sref{mt:past}
                           & \textbf{0.819}~\sblk & 0.806~\sref{mt:luh}  & 0.794~\sref{mt:past} \\
HeLa\textsubscript{DIC}    & 0.960~\sref{mt:siat} & \textbf{0.943}~\sblk & 0.936~\sref{mt:epfl}
                           & \textbf{0.991}~\sblk & 0.989~\sref{mt:siat} & 0.989~\sref{mt:rwth}
                           & 0.931~\sref{mt:siat} & \textbf{0.895}~\sblk & 0.885~\sref{mt:epfl} \\
MSC                        & 0.837~\sref{mt:past} & \textbf{0.825}~\sblk & 0.789~\sref{mt:epfl}
                           & 0.947~\sref{mt:past} & \textbf{0.936}~\sblk & 0.924~\sref{mt:mon}
                           & 0.727~\sref{mt:past} & \textbf{0.713}~\sblk & 0.657~\sref{mt:epfl} \\
A549                       & \textbf{1.000}~\sblk & 1.000~\sblk          & 1.000~\sblk
                           & \textbf{1.000}~\sblk & 1.000~\sblk          & 1.000~\sblk
                           & \textbf{1.000}~\sblk & 1.000~\sblk          & 1.000~\sblk \\
H157                       & \textbf{1.000}~\sblk & 0.993~\sref{mt:past} & 0.972~\sref{mt:epfl}
                           & \textbf{1.000}~\sblk & 0.985~\sref{mt:past} & 0.983~\sref{mt:siat}
                           & \textbf{1.000}~\sblk & 1.000~\sref{mt:past} & 0.962~\sref{mt:epfl} \\
MDA231                     & \textbf{0.929}~\sblk & 0.921~\sref{mt:epfl} & 0.916~\sref{mt:past}
                           & 0.958~\sref{mt:epfl} & \textbf{0.957}~\sblk & 0.957~\sref{mt:past}
                           & \textbf{0.902}~\sblk & 0.883~\sref{mt:epfl} & 0.875~\sref{mt:past} \\
GOWT1                      & 0.887~\sref{mt:past} & \textbf{0.882}~\sblk & 0.849~\sref{mt:kth}
                           & 0.986~\sref{mt:past} & 0.982~\sref{mt:rwth} & \textbf{0.982}~\sblk
                           & 0.788~\sref{mt:past} & \textbf{0.783}~\sblk & 0.723~\sref{mt:kth} \\
HeLa\textsubscript{Fluo}   & \textbf{0.978}~\sblk & 0.974~\sref{mt:past} & 0.960~\sref{mt:kth}
                           & \textbf{0.997}~\sblk & 0.996~\sref{mt:past} & 0.995~\sref{mt:epfl}
                           & \textbf{0.959}~\sblk & 0.951~\sref{mt:past} & 0.926~\sref{mt:kth} \\
CE                         & \textbf{0.950}~\sblk & 0.922~\sref{mt:past} & 0.877~\sref{mt:epfl}
                           & \textbf{0.986}~\sblk & 0.982~\sref{mt:past} & 0.971~\sref{mt:epfl}
                           & \textbf{0.913}~\sblk & 0.862~\sref{mt:past} & 0.782~\sref{mt:epfl} \\
CHO                        & 0.974~\sref{mt:epfl} & \textbf{0.962}~\sblk & 0.953~\sref{mt:rwth}
                           & \textbf{0.993}~\sblk & 0.993~\sref{mt:epfl} & 0.990~\sref{mt:rwth}
                           & 0.955~\sref{mt:epfl} & \textbf{0.931}~\sblk & 0.915~\sref{mt:rwth} \\
U373                       & 0.979~\sref{mt:mon}  & 0.973~\sref{mt:uvic} & 0.972~\sblk
                           & 0.996~\sref{mt:uvic} & 0.995~\sblk          & 0.995~\sblk
                           & 0.964~\sref{mt:mon}  & 0.949~\sblk          & 0.949~\sblk \\
PSC                        & \textbf{0.948}~\sblk & 0.927~\sref{mt:past} & 0.901~\sref{mt:epfl}
                           & \textbf{0.994}~\sblk & 0.992~\sref{mt:past} & 0.991~\sref{mt:epfl}
                           & \textbf{0.902}~\sblk & 0.862~\sref{mt:past} & 0.812~\sref{mt:epfl} \\
A549\textsubscript{SIM}    & \textbf{1.000}~\sblk & 1.000~\sblk          & 1.000~\sblk
                           & \textbf{1.000}~\sblk & 1.000~\sblk          & 1.000~\sblk
                           & \textbf{1.000}~\sblk & 1.000~\sblk          & 1.000~\sblk \\
Fluo2D\textsubscript{SIM+} & 0.989~\sref{mt:siat} & 0.988~\sref{mt:epfl} & 0.983~\sref{mt:past}
                           & 0.999~\sref{mt:epfl} & 0.999~\sref{mt:siat} & 0.999~\sref{mt:uvic}
                           & 0.980~\sref{mt:siat} & 0.977~\sref{mt:epfl} & 0.968~\sref{mt:past} \\
Fluo3D\textsubscript{SIM+} & 0.997~\sref{mt:siat} & \textbf{0.993}~\sblk & 0.992~\sref{mt:past}
                           & 1.000~\sref{mt:siat} & \textbf{0.999}~\sblk & 0.999~\sref{mt:siat}
                           & 0.994~\sref{mt:siat} & \textbf{0.988}~\sblk & 0.984~\sref{mt:siat} \\
\bottomrule
\end{tabular}
\end{table}

\subsection{Bacteria tracking benchmark}
\label{sec:bacteria}

We evaluate on the bacteria tracking benchmark from~\cite{Gallusser:2024:Trackastra}, which features frequent cell divisions in a dense monolayer of rod-shaped bacteria.
Table~\ref{tab:bacteria} compares HOCT against Trackastra with various feature extractors, including powerful pre-trained visual encoders (SAM2.1, DinoV2, microSAM), in both \emph{general} (trained on diverse data) and \emph{specialized} (trained on bacteria) settings.

HOCT achieves the lowest AOGM ($6.36 \pm 1.35$) using only 19-dimensional hand-crafted input features, outperforming the best Trackastra variant (general, SAM2.1: $6.53 \pm 1.68$) which requires a large vision foundation model.
With equivalent hand-crafted features, the specialized Trackastra scores $11.25 \pm 3.67$.

\begin{table}[ht]
    \caption[Bacteria tracking benchmark.]{\textbf{Bacteria tracking benchmark.}
    HOCT achieves the best AOGM with zero deletion errors using only geometric features, even against Trackastra variants that rely on pre-trained visual encoders (SAM2.1, DinoV2, microSAM). Baselines from~\protect\cite{Achard:2024:TrackastraImgFeatures}.}
    \label{tab:bacteria}
    \centering
    \small
    \begin{tabular}{@{}lllcccc@{}}
    \toprule
    & & \textbf{Model} & \textbf{AOGM}$\downarrow$ & \textbf{AOGM-A}$\downarrow$ & \textbf{AOGM-D}$\downarrow$ & \textbf{Division F1}$\uparrow$ \\
    \midrule
    \multirow{6}{*}{\rotatebox[origin=r]{90}{\textbf{General}}} &
    \multirow{5}{*}{\rotatebox[origin=r]{90}{\textbf{Trackastra}}}
    & SAM2.1        & $6.53 \pm 1.68$  & $6.53 \pm 1.68$ & $0.00$ & $\mathbf{0.9981 \pm 0.0005}$ \\
    & & CoTracker3  & $17.67 \pm 7.08$ & $16.56 \pm 6.33$ & $1.11 \pm 1.92$ & $0.9914 \pm 0.0054$ \\
    \cmidrule{3-7}
    & & \textit{Public weights}~\cite{Gallusser:2024:TrackastraCode} & 49.58 & 37.92 & 11.67 & 0.9752 \\
    & & features    & $12.31 \pm 5.43$ & $11.19 \pm 4.46$ & $1.11 \pm 1.92$ & $0.9957 \pm 0.0019$ \\
    & & features-DO & $13.67 \pm 4.86$ & $13.67 \pm 4.86$ & $0.00$ & $0.9945 \pm 0.0022$ \\
    \midrule
    \multirow{9}{*}{\rotatebox[origin=r]{90}{\textbf{Specialized}}} &
    \multirow{8}{*}{\rotatebox[origin=r]{90}{\textbf{Trackastra}}}
    & CoTracker3 & $20.81 \pm 5.41$ & $15.81 \pm 2.93$ & $5.00 \pm 3.59$ & $0.9899 \pm 0.0026$ \\
    & & microSAM   & $11.92 \pm 4.06$ & $10.81 \pm 3.21$ & $1.11 \pm 0.96$ & $0.9960 \pm 0.0013$ \\
    & & DinoV2     & $14.31 \pm 8.02$ & $12.64 \pm 8.02$ & 1.67 & $0.9938 \pm 0.0047$ \\
    & & SAM        & $14.81 \pm 5.69$ & $12.03 \pm 2.41$ & $2.78 \pm 3.85$ & $0.9958 \pm 0.0014$ \\
    & & SAM2.1     & $7.00 \pm 2.14$  & $6.44 \pm 1.78$ & $0.56 \pm 0.96$ & $0.9977 \pm 0.0010$ \\
    \cmidrule{3-7}
    & & \textit{Original (greedy)~\cite{Gallusser:2024:Trackastra}} & 29.0 & -- & -- & -- \\
    & & \textit{Original (ILP)~\cite{Gallusser:2024:Trackastra}} & 19.0 & -- & -- & -- \\
    & & features   & $21.97 \pm 8.53$ & $18.64 \pm 6.84$ & $3.33 \pm 4.55$ & $0.9913 \pm 0.0057$ \\
    & & features-DO   & $11.25 \pm 3.67$ & $9.03 \pm 2.75$ & $2.22 \pm 1.92$ & $0.9963 \pm 0.0019$ \\
    \cmidrule{2-7}
    & \multicolumn{2}{c}{\textbf{HOCT}} & $\mathbf{6.36 \pm 1.35}$ & $\mathbf{6.36 \pm 1.35}$ & $\mathbf{0.00}$ & $0.9962 \pm 0.0012$ \\
    \bottomrule
    \end{tabular}
\end{table}

\subsection{Incremental correction}
\label{sec:correction}

As a final experiment, to assess the argument that edge-embedding is a better paradigma, we evaluate how efficiently each system can be improved with limited user feedback~\cite{Pop:2025:NeedlesInTheHaystack} in an \emph{incremental correction} setting on the three bacteria validation sequences from Sec.~\ref{sec:bacteria}.
We compare against Attrackt~\cite{Lalit:2025:UnsupInteractiveCellTracking}, which builds upon Trackastra, proposing an unsupervised pre-training for tracking and human-in-the-loop fine-tuning.
Both HOCT and Trackastra~+~Attrackt~\cite{Gallusser:2024:Trackastra, Lalit:2025:UnsupInteractiveCellTracking} start from their respective CTC-pretrained weights (never trained on bacteria training data) and run 20 rounds of 20 annotations each (400 total).
HOCT annotates candidate \emph{edges} and refits a logistic regression head on frozen edge embeddings; Trackastra annotates \emph{nodes} and, in its strongest variants developed exclusively for human-in-the-loop training~\cite{Lalit:2025:Attrackt}, it fine-tunes the network via LoRA~\cite{Hu:2022:LoRA}.
A linear classifier head cannot be applied to Trackastra, because it encodes associations as dot-product similarity between node embeddings; the closest analogue is a \emph{linear probe} that learns a single map $L$ applied to both embeddings before the dot product, replacing $\langle \mathbf{n}_i, \mathbf{n}_j \rangle$ with $\langle L\mathbf{n}_i, L\mathbf{n}_j\rangle$, which we also include in the comparison.
Each system uses multiple sampling strategies (Table~\ref{tab:correction}); full details are in Appendix~\ref{app:correction}.

\begin{figure}[ht]
    \centering
    \includegraphics[width=\textwidth]{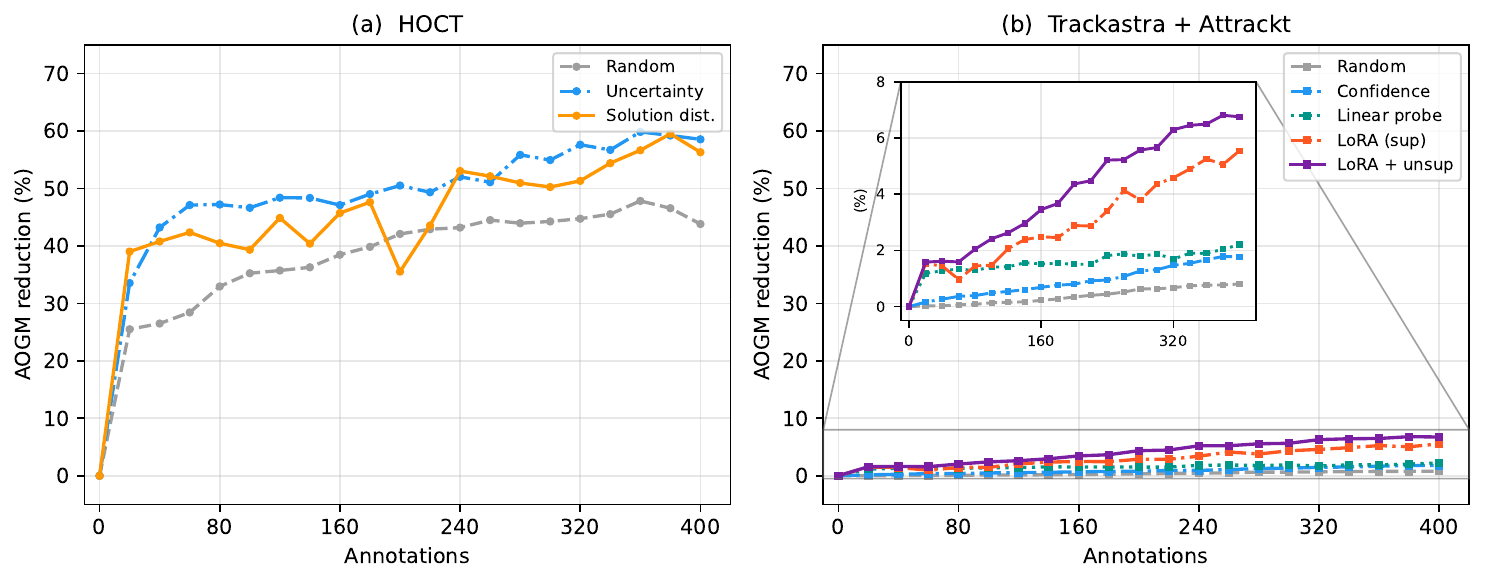}
    \caption{\textbf{Relative AOGM reduction during incremental correction on bacteria validation sequences.}
    Both panels show the percentage AOGM reduction from each system's round-0 baseline, enabling direct comparison of correction efficiency despite different starting points (HOCT: ${\sim}1400$; Trackastra: ${\sim}4950$).
    (a)~HOCT strategies reduce AOGM by up to 59\% within 400 annotations, with \emph{uncertainty} sampling achieving the steepest gains.
    (b)~Trackastra~+~Attrackt~\cite{Lalit:2025:UnsupInteractiveCellTracking} strategies achieve at most $6.75\%$ reduction despite LoRA fine-tuning; the inset zooms into the 0--8\% range to resolve strategy differences.
    \emph{LoRA + unsup} is the strongest Attrackt variant once its image autoencoder embeddings and \texttt{MLP\_E}/\texttt{MLP\_D} weights are pretrained on the validation data (Appendix~\ref{app:correction}); without that pretraining, the unsupervised reconstruction loss is numerically dominant and degrades performance.
    The \emph{linear probe} (a single learned matrix applied to both embeddings before the dot product) reaches $-2.2\%$, slightly above the frozen-model baselines but well below LoRA fine-tuning.
    }
    \label{fig:correction}
\end{figure}

\textbf{Results.}
Table~\ref{tab:correction} and Fig.~\ref{fig:correction} compare both systems across all strategies.
HOCT starts from a better baseline (AOGM $1418$ vs.\ $4951$) and improves much more rapidly, which is unexpected as there is more room for improvement for Trackastra. The best HOCT strategy (\emph{uncertainty}) reduces AOGM by $59\%$ after 400 annotations, while the best Trackastra+Attrackt strategy (\emph{LoRA + unsup}) achieves only a $6.75\%$ reduction despite using LoRA fine-tuning with 2000 gradient steps per round.
The full HOCT round (logistic refit + inference + ILP + metrics on all three sequences) takes ${\sim}$2\,min, ILP-dominated and comparable to Trackastra's frozen-model strategies -- well within the latency budget for active learning, in contrast to the ${\sim}$40--55\,min per round required by LoRA fine-tuning.

\begin{table}[ht]
  \caption{Incremental correction on the bacteria validation sequences.
  Both systems annotate 20 elements per round for 20 rounds (400 total).
  AOGM is averaged over three sequences (lower is better).
  $\Delta$ denotes relative improvement from round 0.
  HOCT uses a frozen backbone with a refitted linear head; Trackastra strategies use LoRA fine-tuning, a learned linear map of the embeddings, and/or ILP constraint pinning. \emph{LoRA + unsup} requires pretrained per-detection autoencoder embeddings and \texttt{MLP\_E}/\texttt{MLP\_D} weights (Appendix~\ref{app:correction}); we pretrain both directly on the validation sequences before round 0.
  }
  \label{tab:correction}
  \centering
  \setlength{\tabcolsep}{4pt}
  \begin{tabular}{@{}llccc@{}}
    \toprule
    System & Strategy & AOGM$_0$ & AOGM$_{20}$ & $\Delta$\,(\%) \\
    \midrule
    \multirow{3}{*}{HOCT}
    & Random              & 1418 & 796  & $-43.8$ \\
    & \textbf{Uncertainty}         & \textbf{1418} & \textbf{587}  & $\mathbf{-58.6}$ \\
    & Solution distance   & 1418 & 619  & $-56.3$ \\
    \midrule
    \multirow{5}{*}{\shortstack[l]{Trackastra\\+~\cite{Lalit:2025:UnsupInteractiveCellTracking}}}
    & Random              & 4951 & 4911 & $-0.8$ \\
    & Confidence (ILP pin)& 4951 & 4863 & $-1.8$ \\
    & Linear probe        & 4951 & 4842 & $-2.2$ \\
    & LoRA (sup)          & 4951 & 4677 & $-5.5$ \\
    & \textbf{LoRA + unsup} & \textbf{4951} & \textbf{4617} & $\mathbf{-6.75}$ \\
    \bottomrule
  \end{tabular}
\end{table}

Because the two systems start from very different baselines (HOCT: 1418; Trackastra: 4951), Fig.~\ref{fig:correction} normalizes each curve to its round-0 value.
HOCT's gains concentrate in the early rounds: the first 80 annotations account for roughly half the total reduction, and active strategies outperform random sampling.
Trackastra~+~Attrackt curves remain near-flat: even LoRA~+~unsup, which updates ${\sim}5\%$ of model weights per round, reinforcing that explicit per-edge features enable cheaper and more effective correction than implicit node-similarity representations.
% The Attrackt unsupervised reconstruction loss (\emph{LoRA + unsup}, $-6.75\%$) outperforms the supervised-only variant (\emph{LoRA (sup)}, $-5.5\%$) once we pretrain the per-detection autoencoder and \texttt{MLP\_E}/\texttt{MLP\_D} weights directly on the validation sequences (Appendix~\ref{app:correction}); without that pretraining, the unsupervised term is numerically dominant and degrades performance.

\section{Conclusion}
\label{sec:conclusion}

HOCT is an edge-centric Transformer that classifies candidate cell links via attention biased by inter-link geometry, addressing both the connected-manifold problem at divisions and the absence of topological label information in candidate tracking graphs. With only simple geometric features and no pre-trained image encoder, it matches or surpasses prior methods across the Cell Tracking Challenge and a bacteria-division benchmark; a logistic head on frozen HOCT features further outperforms LoRA fine-tuning of a Trackastra baseline in a 400-annotation human-in-the-loop setting.

\textbf{Limitations.} The edge-centric architecture comes at increased computational cost.
While standard node-association Transformers have attention complexity $O(|V|^2)$, HOCT requires $O(|E|^2)$ for the edge attention stage.
We mitigate this by constraining the candidate graph to $k$ nearest neighbors per node, yielding $O(|V|^2 k^2)$. This is a significant increase in memory requirements (\eg $100{\times}$ for $k{=}10$), but manageable because the model is compact ($C{=}288$, $2 \times 4$ layers) compared to architectures in other domains.
The spatial distance mask further reduces the effective number of attended pairs.
Extending to very large fields of view with thousands of cells per frame requires tiling, as is currently done when applying image-based models to large microscopy datasets.

\textbf{Broader impact.} Accurate cell tracking advances biological research by enabling quantitative analysis of cell behavior in development, regeneration, and disease models.
Potential downstream applications include drug screening, cancer-metastasis assays, and developmental-biology studies.
As with other biological tools, there is a risk of misuse in unethical research contexts, though we expect the primary impact to be beneficial.

\hfill

\bibliographystyle{plainnat}
\bibliography{references}

%%%%%%%%%%%%%%%%%%%%%%%%%%%%%%%%%%%%%%%%%%%%%%%%%%%%%%%%%%%%

\appendix

\section{Implementation details}
\label{app:implementation}

\textbf{Architecture.}
The full architecture specifications are: input dimension $d{=}19$, hidden dimension $C{=}288$, $H{=}4$ attention heads (head dimension $D{=}72$), MLP expansion factor 2 (hidden MLP dimension 576), attention dropout 0.05, MLP dropout 0.0, gated attention~\cite{Qiu:2025:GatedAttention} in all layers.
The node attention stage uses $L_n{=}4$ layers without distance biases (only RoPE and spatial masking).
The edge attention stage uses $L_e{=}4$ layers with both attractive and repulsive distance biases ($\sigma_h \in \{+1, -1\}$), with independent bias parameters per head.
The distance bias scalar $\alpha_h$ is initialized to $-5$ per head (so $\softplus(-5) \approx 0.007$, starting near zero).
The edge gatherer uses concatenation mode: $\MLP_{\mathrm{gather}}: \mathbb{R}^{2C} \to \mathbb{R}^C$ (two linear layers with GELU). The first edge-attention layer performs cross-attention with raw edge features $\mathbf{f}_e = \mathbf{W}_{\mathrm{edge}}([\mathbf{x}_i \| \mathbf{x}_j])$ as queries and $\mathbf{h}_e^{(0)}$ as keys/values, grounding edge representations in raw feature relationships; subsequent layers perform self-attention among edge tokens.
The classification head is LayerNorm($C$) $\to$ Linear($C$, 1).

\textbf{Input features.}
Each node $i$ carries a $d{=}19$-dimensional feature vector $\mathbf{x}_i$ derived from the segmentation mask: spatiotemporal position $(t, z, y, x)$, equivalent diameter, intensity statistics (min, max, mean, standard deviation), the $3 \times 3$ inertia tensor (9 values), and the distance to the nearest field-of-view border.
All features are standardized per dataset.

\textbf{RoPE configuration.}
Each RoPE instance has learnable frequencies and learnable Householder reflection vectors, with independent parameters per head.
With $D{=}72$ and $d_{\mathrm{pos}}{=}3$, each head has $D/(2 \cdot d_{\mathrm{pos}}) = 12$ frequency bands.

\textbf{Optimizer.}
We use a hybrid Muon--Adam optimizer.
Muon (MomentUm Orthogonalized by Newton-Schulz)~\cite{Jordan:2024:MuonOptimizer} is used for 2D weight matrices in hidden layers (lr=0.015, momentum=0.85).
Adam is applied to biases, norms, input/output projections (lr=0.0017), and positional encoding parameters (lr=0.007).
Weight decay is $1.5 \times 10^{-5}$.
The learning rate schedule is cosine annealing from 100 warmup steps to 50k total steps, with minimum lr $5.5 \times 10^{-5}$.
EMA with decay 0.98 is maintained throughout training.

\textbf{Data augmentation.}
Training augmentations: random spatial crop (256$\times$512$\times$512), 3D affine (rotation $\pm$180\textdegree, scale 0.75--1.25, shear $\pm$0.1), random offset ($\pm$3), axis flips ($p{=}0.5$), intensity power augmentation (exponent 0.75--1.25), and feature dropout ($p{=}0.2$ per feature group).
Validation uses only spatial cropping and standardization.

\textbf{Datasets.}
Training uses 16 CTC datasets with per-dataset sampling weights (range 1.0--4.0) to balance dataset sizes.
Training uses sequence 01, validation uses sequence 02 for each dataset.
Datasets span brightfield (BF-C2DL-HSC, BF-C2DL-MuSC), DIC (DIC-C2DH-HeLa), fluorescence (Fluo-C2DL-MSC, Fluo-C3DH-A549, Fluo-C3DH-A549-SIM, Fluo-C3DH-H157, Fluo-C3DL-MDA231, Fluo-N2DH-GOWT1, Fluo-N2DH-SIM+, Fluo-N2DL-HeLa, Fluo-N3DH-CE, Fluo-N3DH-CHO, Fluo-N3DH-SIM+), and phase contrast (PhC-C2DH-U373, PhC-C2DL-PSC) microscopy.

\textbf{Loss.}
Focal loss with $\gamma{=}3.5$, no positive class reweighting ($\mathrm{pos\_weight}{=}1.0$), division weight $3.5\times$, global averaging (not per-batch).

\textbf{ILP configuration.}
Temporal decay $\lambda{=}0.5$, appearance weight 0.5, disappearance weight 0.25, division weight 0.25, node weight $-10.0$ (strong incentive to include nodes), test-time augmentation with 6 random augmentations. The ILP parameters were found by using half of the datasets from CTC for training and the other half for validation.

\textbf{Compute resources.}
HOCT models were trained on a single NVIDIA H200 GPU node (1 GPU, 15 CPU cores, 128\,GB RAM); one full run (50k steps over all 16 datasets jointly) takes ${\sim}$18\,h with peak GPU memory ${\sim}$60\,GB -- this is the cost per cross-validation split or per ablation row, not per dataset.
The incremental correction experiments (Sec.~\ref{sec:correction}) were run on a desktop with two NVIDIA RTX 3090 GPUs (16 CPU cores, 128\,GB RAM); per-round wall-clocks are listed in Appendix~\ref{app:correction}.

\textbf{Line-to-line distance.}
The minimum distance between two finite line segments in $\mathbb{R}^n$ (Eq.~\ref{eq:line_dist}) admits an analytical solution.
Each segment $e_m = (i_m, j_m)$ is parameterized as $\mathbf{p}_{i_m} + s\,\mathbf{d}_m$ with $s \in [0,1]$ and $\mathbf{d}_m = \mathbf{p}_{j_m} - \mathbf{p}_{i_m}$.
For a pair of segments $e_m, e_n$, let $\mathbf{v} = \mathbf{p}_{i_m} - \mathbf{p}_{i_n}$ and define:
\begin{align}
    a = \mathbf{d}_m \!\cdot\! \mathbf{d}_m,\;\;
    b = \mathbf{d}_m \!\cdot\! \mathbf{d}_n,\;\;
    c = \mathbf{d}_n \!\cdot\! \mathbf{d}_n,\;\;
    f = \mathbf{d}_m \!\cdot\! \mathbf{v},\;\;
    g = \mathbf{d}_n \!\cdot\! \mathbf{v}
\end{align}
Minimizing $\|\mathbf{v} + s\,\mathbf{d}_m - u\,\mathbf{d}_n\|^2$ over unconstrained $s, u$ yields the linear system:
\begin{align}
    \begin{pmatrix} a & -b \\ -b & c \end{pmatrix}
    \begin{pmatrix} s \\ u \end{pmatrix}
    = \begin{pmatrix} -f \\ g \end{pmatrix}
    \label{eq:seg_system}
\end{align}
For non-parallel segments ($ac - b^2 > 0$), the unconstrained solution is:
\begin{align}
    s^* = \frac{bg - cf}{ac - b^2}, \qquad u^* = \frac{bs^* + g}{c}
    \label{eq:seg_unclamped}
\end{align}
Since $s, u$ must lie in $[0,1]$ for finite segments, we apply iterative projection:
\begin{enumerate}[nosep]
    \item $s_0 = \mathrm{clamp}(s^*,\; 0,\; 1)$,
    \item $u_0 = \mathrm{clamp}\!\bigl(\tfrac{b\,s_0 + g}{c},\; 0,\; 1\bigr)$,
    \item $s_1 = \mathrm{clamp}\!\bigl(\tfrac{b\,u_0 - f}{a},\; 0,\; 1\bigr)$.
\end{enumerate}
For nearly parallel segments ($ac - b^2 < \epsilon$), we fix $s_0 = 0$ and set $u_0 = \mathrm{clamp}(g/c,\; 0,\; 1)$.
The line-to-line distance is then:
\begin{align}
    d_{\mathrm{line}}(e_m, e_n) = \bigl\|(\mathbf{p}_{i_m} + s_1\,\mathbf{d}_m) - (\mathbf{p}_{i_n} + u_0\,\mathbf{d}_n)\bigr\|
    \label{eq:line_dist_solution}
\end{align}
The computation is fully vectorized, computed only once per forward pass, and detached from the autograd graph (distances are fixed geometric constants per batch).

\section{ILP formulation}
\label{app:ilp}

We formulate tracking as a binary linear program adapting the ILP formulation from~\cite{Zhang:2008:MultiObjTrackingNetworkFlow,Turetken:2016:NetworkFlow}.
Candidate edges are directed forward in time, $(i,j) \in E_C \Rightarrow t_i < t_j$ (Sec.~\ref{sec:method}).
For each node $i \in V_C$, we define binary variables: $y_i$ (node is active), $a_i$ (appearance, starts a new track), $b_i$ (disappearance, ends a track), and $\delta_i$ (division, node has two daughters).
For each candidate edge $e \in E_C$, we define a binary variable $x_e$.
The ILP minimizes the total weighted score:
\begin{align}
    \min_{\mathbf{x}, \mathbf{y}} \quad \sum_{e \in E_C} w_e \, x_e + \sum_{i \in V_C} \bigl(w_n \, y_i + w_a(i) \, a_i + w_b \, b_i + w_\delta \, \delta_i \bigr) \label{eq:ilp_obj}
\end{align}
subject to flow conservation at each node:
\begin{align}
    y_j &= a_j + \sum_{i :\, (i,j) \in E_C} x_{(i,j)} && \forall \, j \in V_C \label{eq:ilp_inflow} \\
    y_i + \delta_i &= b_i + \sum_{j :\, (i,j) \in E_C} x_{(i,j)} && \forall \, i \in V_C \label{eq:ilp_outflow} \\
    y_i &\geq \delta_i && \forall \, i \in V_C \label{eq:ilp_div}
\end{align}
with $y_i, a_i, b_i, \delta_i \in \{0,1\}$ for all $i \in V_C$ and $x_e \in \{0,1\}$ for all $e \in E_C$.
Eq.~\eqref{eq:ilp_inflow} ensures each active node either appears or has exactly one incoming edge (parent).
Eq.~\eqref{eq:ilp_outflow} balances outflow: an active node either disappears or produces outgoing edges, with division allowing two.
Eq.~\eqref{eq:ilp_div} ensures only active nodes can divide.

Edge weights $w_e$ are derived from the predicted parental softmax probabilities and damped by the temporal gap,
\begin{align}
    w_e = (-p_e + 0.5)\,\exp\!\bigl(-\lambda(\Delta t_e - 1)\bigr)
\end{align}
so that high-confidence edges receive a negative cost (favoring selection) and longer-range candidates are progressively penalized; $\lambda$ is the same temporal-decay constant used in Eq.~\eqref{eq:appearance}.
The node incentive $w_n < 0$ rewards keeping detections, so the slack variables ($a_i$, $b_i$, $\delta_i$) are activated only when no edge configuration explains the node.
In the standard formulation, the slack costs ($w_a$, $w_b$, $w_\delta$) are fixed non-negative constants.
We replace the appearance constant with a variable per-node cost
\begin{align}
    w_a(i) = \bar w_a \,\bigl(1 - p_\alpha(i)\bigr)
\end{align}
where $\bar w_a$ is the base appearance weight and $t_{\min}$ is the first frame (Sec.~\ref{sec:ilp}): cells the model confidently identifies as appearing pay no penalty ($w_a(i) = 0$ when $p_\alpha(i) = 1$), and detections in the first frame appear at zero cost.
Numerical values for $w_n$, $\bar w_a$, $w_b$, $w_\delta$, and $\lambda$ are listed in Appendix~\ref{app:implementation}.

\section{Edge heterophily analysis}
\label{app:heterophily}

We quantify the lack of topological label information in candidate tracking graphs by analyzing their line graphs.
Given a candidate graph $G_C = (V_C, E_C)$, we construct the line graph $L(G_C)$: each edge $e \in E_C$ becomes a node in $L(G_C)$, and two nodes in $L(G_C)$ are connected whenever the corresponding edges in $G_C$ share an endpoint.
Each line-graph node inherits the binary ground-truth label of its original edge (positive if the link is in the GT lineage, negative otherwise).
Let $E_L$ denote the edge set of $L(G_C)$.

We report three metrics:

\textbf{Edge homophily ratio}~\cite{Zheng:2022:GraphHeterophilySurvey}.
The fraction of line-graph edges whose two endpoints carry the same label:
\begin{align}
    \mathcal{H}_{\mathrm{edge}} = \frac{|\{(e_m, e_n) \in E_L : y_{e_m} = y_{e_n}\}|}{|E_L|}
\end{align}
This metric is sensitive to class imbalance: when most edges are negative (as in tracking, with positive rate ${\sim}20\%$), negative--negative agreement inflates $\mathcal{H}_{\mathrm{edge}}$ even under random labeling.

\textbf{Adjusted homophily}~\cite{Platonov:2023:CriticalLookHeterophily}.
Corrects $\mathcal{H}_{\mathrm{edge}}$ by subtracting its expected value under random label assignment:
\begin{align}
    \mathcal{H}_{\mathrm{adj}} = \frac{\mathcal{H}_{\mathrm{edge}} - \sum_{c} (D_c / 2|E_L|)^2}{1 - \sum_{c} (D_c / 2|E_L|)^2}
\end{align}
where $D_c$ is the sum of degrees of line-graph nodes with label $c$.
$\mathcal{H}_{\mathrm{adj}} = 0$ indicates random-level agreement; positive values indicate homophily; negative values indicate heterophily.

\textbf{Positive-neighbor ratio.}
For each true (GT) edge, the fraction of its co-incident edges that are also true.
This directly measures how isolated correct links are among their graph neighbors.

\textbf{Results.}
Across the 28 non-degenerate CTC sequences (excluding 4 sequences where tracking is trivial and every candidate edge is correct), we obtain:
$\mathcal{H}_{\mathrm{edge}} = 0.645 \pm 0.051$,
$\mathcal{H}_{\mathrm{adj}} = 0.012 \pm 0.044$,
positive-neighbor ratio $= 0.29 \pm 0.15$.
The near-zero adjusted homophily confirms that graph topology carries zero label information: knowing the label of one edge tells you nothing about its topological neighbors.
For each true link, $71\%$ of its graph neighbors carry a conflicting label, meaning that topology-driven aggregation (whether via GNN message passing or RoPE-biased attention) dilutes the signal of correct associations.
For reference, standard homophilic benchmarks (Cora, CiteSeer) have $\mathcal{H}_{\mathrm{adj}} > 0.5$~\cite{Platonov:2023:CriticalLookHeterophily}, where message passing is effective because topology is informative.
Candidate tracking graphs, at $\mathcal{H}_{\mathrm{adj}} \approx 0$, are the worst case for any method that relies on graph structure for label prediction.

\section{Incremental correction baseline details}
\label{app:correction}

\textbf{HOCT correction mechanism.}
The HOCT backbone is kept frozen throughout correction.
At each round, a single-layer logistic regression head ($\mathbf{w}^\top \mathbf{x} + b$) is fitted from scratch on the edge-level features extracted by the backbone for all accumulated labeled edges.
The head is always re-initialized from the backbone's original classification layer (no warm-start across rounds).
Training uses full-batch L-BFGS (500 iterations) with class-balanced binary cross-entropy ($\mathrm{pos\_weight} = n_{\mathrm{neg}} / n_{\mathrm{pos}}$) and two regularization terms:
(i)~\emph{adaptive L2}, $\lambda_{\ell_2} \cdot \frac{d}{n} \|\mathbf{w}\|^2$, which penalizes large head weights but weakens as more labels $n$ accumulate ($d$ is the feature dimension, $\lambda_{\ell_2}{=}1$);
(ii)~\emph{consistency loss}, $\lambda_c \cdot \frac{d}{n} \, \mathrm{BCE}(\hat{y}_u, y_u^{\mathrm{ILP}})$, which anchors predictions on \emph{unlabeled} edges to the current ILP solution, preventing local corrections from propagating destructively ($\lambda_c{=}0.25$).
Features are extracted with 5 test-time augmentations (overlapping temporal windows).

The three sampling strategies select which 20 edges to label each round:
\emph{random} samples uniformly (balanced across positive/negative ILP labels);
\emph{uncertainty} selects edges closest to the 0.5 decision boundary;
\emph{solution distance} selects edges with the largest disagreement between the ILP solution and the predicted similarity, restricted to the ambiguity band $[0.15, 0.85]$.

\textbf{Trackastra~+~Attrackt correction mechanism.}
The Trackastra pipeline produces per-window association matrices $A(v)$ giving the predicted probability that each candidate edge is active; like ours, an ILP solver is used to select a globally consistent tracking solution.
At each correction round, the oracle annotates 20 \emph{nodes} (cell detections) and provides their ground-truth parent link(s).
These annotated edges are injected as hard constraints into the ILP, forcing the solver to match the user annotations regardless of the predicted association scores.
\emph{Random} selects nodes uniformly at random.
\emph{Confidence} ranks nodes by their disagreement score $\lvert \sum A(v) - \sum A_{\mathrm{solver}}(v) \rvert$, i.e.\ the absolute difference between the sum of predicted association weights and the number of edges selected by the ILP for that node; nodes with the highest disagreement are annotated first.
Both strategies are a baseline and update the tracking only through ILP constraint pinning, without modifying the Trackastra model weights.
\emph{Linear probe} keeps the Trackastra backbone frozen and learns a single shared linear map $L \in \mathbb{R}^{d \times d}$ ($d{=}512$, $\approx 262$k parameters, $1.1\%$ of the model) applied to both query and key embeddings before the outer-product association matrix, so $A_{ij} = \langle L\,\mathrm{head}_x(\mathbf{z}_i),\, L\,\mathrm{head}_y(\mathbf{z}_j)\rangle$. $L$ is initialised to the identity (so round 0 is unchanged), trained with the same supervised + top-10\% pseudo-supervised loss as \emph{LoRA (sup)} (500 gradient steps per round, continuing from the previous round's $L$), and used at inference in place of the identity. This is the closest analogue to HOCT's logistic-head probe within Trackastra's dot-product association: only the embedding-space metric is updated, never the network weights.
\emph{LoRA (sup)} additionally fine-tunes the Trackastra transformer via LoRA~\cite{Hu:2022:LoRA} (rank 32, applied to attention Q/K/V/projection layers) using $\mathcal{L}_{\mathrm{sup}} + \mathcal{L}_{\mathrm{pseudo\text{-}sup}}$ (2000 gradient steps per round, continuing from the previous round's adapter), where the pseudo-supervised loss treats the ILP solution for the top-10\% most confident nodes as soft labels.
\emph{LoRA + unsup} extends this with the full Attrackt three-component loss $\mathcal{L}_{\mathrm{sup}} + \mathcal{L}_{\mathrm{pseudo\text{-}sup}} + \mathcal{L}_{\mathrm{unsup}}$; the unsupervised term is the Attrackt feature-reconstruction loss over all frames, where the autoencoder embeddings and the encoder/decoder MLPs are pretrained on the validation sequences before round 0 (see the dedicated bullet below) and the LoRA adapters then continue training jointly with the MLPs.

\textbf{Trackastra baseline differences from~\cite{Lalit:2025:UnsupInteractiveCellTracking}.}
We reproduce the interactive fine-tuning pipeline with the following differences from the original paper:

\begin{itemize}[noitemsep, topsep=0pt]
  \item \textbf{Starting checkpoint.}
  The original paper trains a Trackastra model from scratch using the Attrackt unsupervised loss, then fine-tunes interactively.
  We instead start from the publicly available \texttt{ctc} pretrained Trackastra checkpoint~\cite{Gallusser:2024:Trackastra}, which was trained on the same CTC datasets with the standard supervised loss.
  This represents the most accessible starting point for a practitioner.

  \item \textbf{Cell embeddings.}
  The Attrackt unsupervised loss requires per-detection embeddings from a frozen image autoencoder. We follow the paper recipe and train an autoencoder on detection-centred crops from the validation sequences, then run inference to obtain a 64-dimensional embedding per detection (full configuration in the \emph{Pretraining for \emph{LoRA + unsup}} bullet below).

  \item \textbf{Loss weighting.}
  Following the original paper, we weight all three fine-tuning loss terms equally ($\mathcal{L}_{\mathrm{sup}} + \mathcal{L}_{\mathrm{pseudo\text{-}sup}} + \mathcal{L}_{\mathrm{unsup}}$) and use the top-10\% most confident nodes as pseudo-labels.

  \item \textbf{Annotation granularity.}
  Each system follows its native annotation protocol: the Trackastra baseline annotates 20 \emph{nodes} per round (as defined by Attrackt), while HOCT annotates 20 \emph{edges} per round.
  Node annotation is arguably easier because each node yields 1--2 ground-truth edges, so the Trackastra baseline receives slightly more edge-level supervision per round.
  This favors the baseline, making the comparison conservative for HOCT.

  \item \textbf{Pretraining for \emph{LoRA + unsup}.}
  The Attrackt unsupervised reconstruction loss (\emph{LoRA + unsup}) requires per-detection autoencoder embeddings as targets and two small MLPs (``MLP\_E''/``MLP\_D'') that wrap them inside the loss. The publicly released \texttt{ctc} Trackastra checkpoint ships neither, so we pretrain both directly on the bacteria validation sequences before round 0:
  (i)~train an image autoencoder (paper recipe: $64\!\times\!64$ crops, four downsample stages, $4\!\times\!4\!\times\!4$ latent flattened to a 64-dim per-detection embedding) on detection-centred crops for 50k iterations; (ii)~run the encoder over every detection in the validation CSV to produce the embeddings file consumed by Attrackt's data loader; (iii)~pretrain ``MLP\_E''/``MLP\_D'' (each $\approx$17k parameters, $d_{\mathrm{model}}{=}64$) against those embeddings using \texttt{common\_unsupervised\_step} for 10k iterations with the \texttt{ctc} backbone frozen, so the round-0 baseline is unchanged. The pretrained MLPs warm-start round 1 of the correction loop; subsequent rounds continue from the previous round's saved MLPs.
  Without this pretraining, the unsupervised loss is initially $\sim$1000$\times$ larger than the supervised one and dominates the gradient (the supervised signal becomes noise); with the pretraining the three losses are within an order of magnitude at the start of each round, and \emph{LoRA + unsup} becomes the strongest Attrackt variant.

  \item \textbf{Computational cost.}
  All times below are full per-round wall clock on the three bacteria validation sequences combined and on the same RTX 3090, including head/adapter fit, inference (with TTA where applicable), motile ILP, and traccuracy evaluation.
  HOCT takes ${\sim}$2\,min per round, dominated by the per-sequence ILP solve; the logistic head refit itself is ${<}$1\,s.
  Trackastra ILP-only strategies (random, confidence) require ${\sim}$3\,min per round, with the same ILP step as the dominant cost (no model update).
  The linear probe takes ${\sim}$10\,min per round (5--6\,min fitting $L$ + 3--4\,min eight-way TTA inference + ILP), roughly $4{\times}$ cheaper than \emph{LoRA (sup)} because the frozen transformer body runs under \texttt{torch.no\_grad} and only $L$ is in the autograd graph.
  LoRA strategies require ${\sim}$40\,min (\emph{LoRA (sup)}) or ${\sim}$55\,min (\emph{LoRA + unsup}) per round, including the same eight-way TTA. \emph{LoRA + unsup} additionally requires a one-off pretraining pass (${\sim}$30\,min for the autoencoder + 10\,min for the MLPs on the same GPU).

  \item \textbf{Code.}
  Our Trackastra~+~Attrackt baseline builds on the publicly available Attrackt library~\cite{Lalit:2025:Attrackt} and experiment code~\cite{Lalit:2025:AttracktExperiments}.
\end{itemize}

\begin{figure}[ht]
    \centering
    \includegraphics[width=\textwidth]{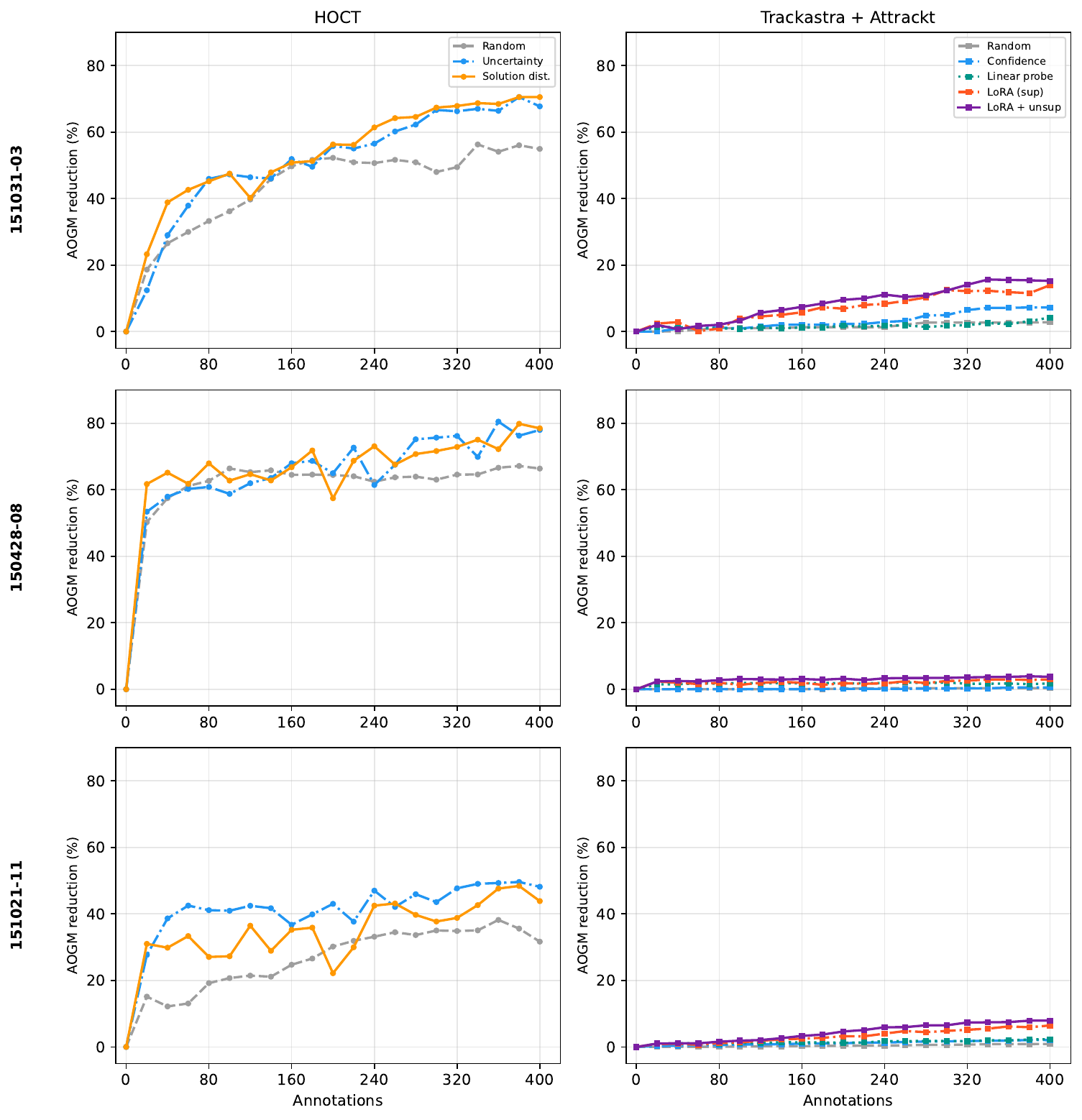}
    \caption{\textbf{Per-sequence AOGM progression during incremental correction.}
    Each row corresponds to one bacteria validation sequence.
    Left column: HOCT strategies; right column: Trackastra~+~Attrackt strategies.
    The improvement percentages are consistent across sequences: HOCT reduces AOGM by 48--78\%, while Trackastra~+~Attrackt achieves between $\sim$4\% (150428-08) and $\sim$15\% (151031-03), with \emph{LoRA + unsup} the strongest variant in every sequence.
    }
    \label{fig:correction_per_seq}
\end{figure}

%%%%%%%%%%%%%%%%%%%%%%%%%%%%%%%%%%%%%%%%%%%%%%%%%%%%%%%%%%%%

\section{Full CTC leaderboard scores and ranks}
\label{app:ctc_full}

Table~\ref{tab:ctc_full} reports the official CTC scores and current leaderboard ranks\footnote{Provided directly by the organizers; the public leaderboard had not yet been refreshed as of 2026-05-01.} of our generalizable HOCT model on the $16$ linking challenge test datasets, extending Table~\ref{tab:ctc} from Section~\ref{sec:ctc} with the entries that fall outside the top three.
Each cell shows the score followed by the rank as SCORE~(RANK), where RANK is HOCT's position among the N currently submitted methods on that dataset/metric.
The \emph{Generalizability} row reports the CTC-official overall-performance score, which aggregates across datasets.
The two datasets where HOCT clearly falls outside the top three are PhC-C2DH-U373 (rank 7--8) and Fluo-N2DH-SIM+ (rank 4--5): both contain few cells and few divisions per frame, so small absolute error counts let dataset-specialist methods pull ahead of a generalist model trained jointly on 16 datasets.

\begin{table}[ht]
\caption{\textbf{Full CTC hidden-test results for HOCT.}
Official scores and current leaderboard ranks, as SCORE~(RANK), of our generalizable HOCT model trained jointly on all 2D and 3D CTC datasets.
}
\label{tab:ctc_full}
\centering
\small
\begin{tabular}{@{}lccc@{}}
\toprule
Dataset & CLB & LNK & BIO \\
\midrule
Generalizability           & 0.930~(1) & 0.985~(1) & 0.875~(1) \\
\midrule
HSC                        & 0.845~(2) & 0.992~(1) & 0.699~(2) \\
MuSC                       & 0.902~(1) & 0.986~(1) & 0.819~(1) \\
HeLa\textsubscript{DIC}    & 0.943~(2) & 0.991~(1) & 0.895~(2) \\
MSC                        & 0.825~(2) & 0.936~(2) & 0.713~(2) \\
A549                       & 1.000~(1) & 1.000~(1) & 1.000~(1) \\
H157                       & 1.000~(1) & 1.000~(1) & 1.000~(1) \\
MDA231                     & 0.929~(1) & 0.957~(2) & 0.902~(1) \\
GOWT1                      & 0.882~(2) & 0.982~(3) & 0.783~(2) \\
HeLa\textsubscript{Fluo}   & 0.978~(1) & 0.997~(1) & 0.959~(1) \\
CE                         & 0.950~(1) & 0.986~(1) & 0.913~(1) \\
CHO                        & 0.962~(2) & 0.993~(1) & 0.931~(2) \\
U373                       & 0.925~(8) & 0.993~(7) & 0.856~(7) \\
PSC                        & 0.948~(1) & 0.994~(1) & 0.902~(1) \\
A549\textsubscript{SIM}    & 1.000~(1) & 1.000~(1) & 1.000~(1) \\
Fluo2D\textsubscript{SIM+} & 0.982~(5) & 0.998~(5) & 0.966~(4) \\
Fluo3D\textsubscript{SIM+} & 0.993~(2) & 0.999~(2) & 0.988~(2) \\
\bottomrule
\end{tabular}
\end{table}

\newpage
\input{checklist}

\end{document}

%% file: checklist.tex
\section*{NeurIPS Paper Checklist}

\begin{enumerate}

\item {\bf Claims}
    \item[] Question: Do the main claims made in the abstract and introduction accurately reflect the paper's contributions and scope?
    \item[] Answer: \answerYes{}
    \item[] Justification: The abstract and introduction (Sec.~\ref{sec:introduction}) list four contributions --- edge-centric Transformer architecture, geometric distance encoding, multi-frame parental softmax with ILP improvements, and incremental correction --- each supported by corresponding experimental sections (Secs.~\ref{sec:ablation_ilp}--\ref{sec:correction}).

\item {\bf Limitations}
    \item[] Question: Does the paper discuss the limitations of the work performed by the authors?
    \item[] Answer: \answerYes{}
    \item[] Justification: Section~\ref{sec:conclusion}, paragraph \textbf{Limitations.} discusses the $O(|E|^2)$ attention complexity, the resulting memory increase (${\sim}100{\times}$ for $k{=}10$ neighbors), and the need for tiling on very large fields of view.

\item {\bf Theory assumptions and proofs}
    \item[] Question: For each theoretical result, does the paper provide the full set of assumptions and a complete (and correct) proof?
    \item[] Answer: \answerNA{}
    \item[] Justification: The paper does not include theoretical results (theorems, lemmas, or formal proofs). The mathematical content consists of model definitions (architecture, loss, ILP formulation) and an analytical formula for segment-to-segment distance (Appendix~\ref{app:implementation}), which is a standard geometric computation.

    \item {\bf Experimental result reproducibility}
    \item[] Question: Does the paper fully disclose all the information needed to reproduce the main experimental results of the paper to the extent that it affects the main claims and/or conclusions of the paper (regardless of whether the code and data are provided or not)?
    \item[] Answer: \answerYes{}
    \item[] Justification: Full architecture specifications, optimizer settings, learning rate schedule, data augmentation pipeline, loss configuration, ILP parameters, input features, and RoPE configuration are provided in Appendix~\ref{app:implementation}. The cross-validation protocol and dataset list are described in Sec.~\ref{sec:results}. The incremental correction protocol is detailed in Appendix~\ref{app:correction}.

\item {\bf Open access to data and code}
    \item[] Question: Does the paper provide open access to the data and code, with sufficient instructions to faithfully reproduce the main experimental results, as described in supplemental material?
    \item[] Answer: \answerYes{}
    \item[] Justification: An anonymized copy of the HOCT source code and trained model weights is included as supplementary material for reviewing purposes. Upon acceptance, the de-anonymized code and weights will be released publicly on GitHub under the BSD-3-Clause license. All datasets and baselines used in the experiments are already publicly available: the Cell Tracking Challenge datasets~\cite{Mavska:2023:CTC, Ulman:2017:ObjectiveCompCTC}, the bacteria tracking benchmark and Trackastra codebase from~\cite{Gallusser:2024:Trackastra}, and the Attrackt code from~\cite{Lalit:2025:UnsupInteractiveCellTracking}. Full architecture, hyperparameters, optimizer schedule, data augmentation, ILP configuration, and the incremental correction protocol are documented in Appendices~\ref{app:implementation} and~\ref{app:correction}.

\item {\bf Experimental setting/details}
    \item[] Question: Does the paper specify all the training and test details (e.g., data splits, hyperparameters, how they were chosen, type of optimizer) necessary to understand the results?
    \item[] Answer: \answerYes{}
    \item[] Justification: Sec.~\ref{sec:results} describes the cross-validation protocol, metrics, and distance threshold. Appendix~\ref{app:implementation} provides all hyperparameters: architecture dimensions, optimizer (Muon--Adam hybrid with per-parameter-group learning rates), learning rate schedule, data augmentation, loss function, ILP configuration, and input features. Appendix~\ref{app:correction} details the incremental correction protocol for both HOCT and the Trackastra baseline.

\item {\bf Experiment statistical significance}
    \item[] Question: Does the paper report error bars suitably and correctly defined or other appropriate information about the statistical significance of the experiments?
    \item[] Answer: \answerYes{}
    \item[] Justification: The bacteria benchmark (Table~\ref{tab:bacteria}) and edge-stage aggregation ablation (Table~\ref{tab:gnn_ablation}) report mean $\pm$ standard deviation across cross-validation folds or sequences. The CTC ablation (Table~\ref{tab:ablation}) reports results for both cross-validation splits individually rather than aggregating, allowing the reader to assess variability. The splits are fixed by the dataset provider and the ILP solution is deterministic.

\item {\bf Experiments compute resources}
    \item[] Question: For each experiment, does the paper provide sufficient information on the computer resources (type of compute workers, memory, time of execution) needed to reproduce the experiments?
    \item[] Answer: \answerYes{}
    \item[] Justification: Appendix~\ref{app:implementation} reports hardware (NVIDIA H200 for training, two NVIDIA RTX 3090s for correction experiments), CPU/RAM specifications, and approximate training time (${\sim}$18\,h per dataset configuration). Appendix~\ref{app:correction} reports per-round wall-clock times for the correction experiment ($<$1\,s for HOCT, ${\sim}$35\,min for LoRA).

\item {\bf Code of ethics}
    \item[] Question: Does the research conducted in the paper conform, in every respect, with the NeurIPS Code of Ethics \url{https://neurips.cc/public/EthicsGuidelines}?
    \item[] Answer: \answerYes{}
    \item[] Justification: The research uses publicly available cell tracking benchmarks, involves no human subjects, and poses no foreseeable ethical concerns.

\item {\bf Broader impacts}
    \item[] Question: Does the paper discuss both potential positive societal impacts and negative societal impacts of the work performed?
    \item[] Answer: \answerYes{}
    \item[] Justification: Section~\ref{sec:conclusion}, paragraph \textbf{Broader impact.} discusses positive applications (drug screening, developmental biology, cancer metastasis research) and acknowledges the risk of misuse in unethical research contexts.

\item {\bf Safeguards}
    \item[] Question: Does the paper describe safeguards that have been put in place for responsible release of data or models that have a high risk for misuse (e.g., pre-trained language models, image generators, or scraped datasets)?
    \item[] Answer: \answerNA{}
    \item[] Justification: The model is a domain-specific cell tracking tool that does not pose risks of misuse comparable to generative models or large language models.

\item {\bf Licenses for existing assets}
    \item[] Question: Are the creators or original owners of assets (e.g., code, data, models), used in the paper, properly credited and are the license and terms of use explicitly mentioned and properly respected?
    \item[] Answer: \answerYes{}
    \item[] Justification: All existing assets used in this work are publicly available and properly cited: the Cell Tracking Challenge datasets~\cite{Mavska:2023:CTC, Ulman:2017:ObjectiveCompCTC}, the bacteria tracking benchmark and Trackastra codebase from~\cite{Gallusser:2024:Trackastra}, and the Attrackt code from~\cite{Lalit:2025:UnsupInteractiveCellTracking}. They are used as baselines and benchmark data under their original publicly available terms.

\item {\bf New assets}
    \item[] Question: Are new assets introduced in the paper well documented and is the documentation provided alongside the assets?
    \item[] Answer: \answerYes{}
    \item[] Justification: The HOCT source code and trained model weights will be released on GitHub under the BSD-3-Clause license upon acceptance, accompanied by documentation covering installation, training, and evaluation, together with the configuration files used to produce the reported results.

\item {\bf Crowdsourcing and research with human subjects}
    \item[] Question: For crowdsourcing experiments and research with human subjects, does the paper include the full text of instructions given to participants and screenshots, if applicable, as well as details about compensation (if any)?
    \item[] Answer: \answerNA{}
    \item[] Justification: The paper does not involve crowdsourcing or research with human subjects. The ``incremental correction'' experiment uses a simulated oracle with access to ground-truth annotations, not human annotators.

\item {\bf Institutional review board (IRB) approvals or equivalent for research with human subjects}
    \item[] Question: Does the paper describe potential risks incurred by study participants, whether such risks were disclosed to the subjects, and whether Institutional Review Board (IRB) approvals (or an equivalent approval/review based on the requirements of your country or institution) were obtained?
    \item[] Answer: \answerNA{}
    \item[] Justification: The paper does not involve research with human subjects.

\item {\bf Declaration of LLM usage}
    \item[] Question: Does the paper describe the usage of LLMs if it is an important, original, or non-standard component of the core methods in this research? Note that if the LLM is used only for writing, editing, or formatting purposes and does \emph{not} impact the core methodology, scientific rigor, or originality of the research, declaration is not required.
    %this research?
    \item[] Answer: \answerNA{}
    \item[] Justification: The core method does not involve LLMs. No LLM is used as a component of the architecture, training, or evaluation pipeline.

\end{enumerate}